\documentclass[10pt, a4paper]{article}

\usepackage{multibib}
\usepackage{multirow}
\usepackage{graphicx}
\usepackage{tabularx}
\usepackage{soul}
\usepackage{epstopdf}
\usepackage[utf8]{inputenc}
\usepackage{hyperref}
\usepackage{xstring}
\usepackage{color}
\usepackage{booktabs}
\usepackage{amsmath}
\usepackage{fancyvrb}
\usepackage{tikz}
\usepackage{pgfplots}
\pgfplotsset{compat=1.18}
\usepackage{subcaption}
\usepackage{float}

\usepackage[final]{lrec2026} 

\title{\textsc{SiPaKosa}: A Comprehensive Corpus of Canonical and Classical Buddhist Texts in Sinhala and Pali}

\name{Ranidu Gurursinghe$^\spadesuit$ and Nevidu Jayatilleke$^\clubsuit$}

\address{$^\spadesuit$School of Computing, Informatics Institute of Technology, Sri Lanka \\
$^\clubsuit$Department of Computer Science \& Engineering, University of Moratuwa, Sri Lanka \\ \texttt{ranidu.20222198@iit.ac.lk}, \texttt{nevidu.25@cse.mrt.ac.lk}}

\abstract{
\textsc{SiPaKosa} is a comprehensive corpus of Sinhala and Pali doctrinal texts comprising approximately 786K sentences and 9.25M words, incorporating 16 copyright-cleared historical Buddhist documents alongside the complete web-scraped Tripiṭaka canonical texts. The corpus was created through high-quality OCR using \texttt{Google Document AI} on historical manuscripts, combined with systematic web scraping of canonical repositories, followed by rigorous quality control and metadata annotation. The corpus is organised into language-specific subcorpora: Sinhala and Mixed Sinhala-Pali. We evaluate the performance of language models using ten pretrained models, with perplexity scores ranging from 1.09 to 189.67 on our corpus. This analysis shows that proprietary models significantly outperform open-source alternatives by factors of three to six times. This corpus supports the pretraining of domain-adapted language models, facilitates historical language analysis, and aids in the development of information retrieval systems for Buddhist scholarship while preserving Sinhala cultural heritage.
\\ \newline }

\begin{document}

\maketitleabstract

\section{Introduction}
\label{sec:introduction}

\raisebox{-0.5ex}{%
    \includegraphics[height=1.4\fontcharht\font`\A]{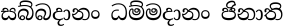}%
}\footnote{\scriptsize \textbf{English Meaning:} \textit{The gift of dhamma (truth) conquers all other gifts; it truly excels and surpasses them all}.\\ \textbf{Sinhala to IPA Transliteration:} \raisebox{-0.5ex}{%
    \includegraphics[height=1.5\fontcharht\font`\A]{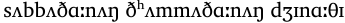}%
}.
}~\cite{thera1954dhammapada}, highlights that sharing knowledge and truth is considered the highest form of generosity because it provides the tools for others to free themselves. While the importance of the universal sharing of knowledge is highlighted, the field of computational linguistics remains significantly under-resourced in terms of structured religious corpora~\cite{hutchinson-2024-modeling}.

The Sinhala language is part of the Indo-Aryan branch of the Indo-European language family with a rich and diverse literary heritage that has developed over several millennia. Its origins can be traced back to between the 3rd and 2nd centuries BCE. It is the primary language of the Sinhalese people, who represent the largest ethnic group in Sri Lanka, and it is recognised as the first language (L1) for approximately 16 million people~\cite{desilva2026survey}. Furthermore, Sinhala is classified as a lower-resourced language (Category 01) according to the criteria presented by~\citet{ranathunga-de-silva-2022-languages}.

Pali, which was historically a language commonly used by monks of different nations for communication~\cite{zigmond-2023-distinguishing}, is considered a dead language despite being widely studied because it is the language of early Buddhist scriptures~\cite{knauth-alfter-2014-dictionary}. In terms of data resources, Pali is also a category 1 low-resource language according to the criteria presented by~\citet{ranathunga-de-silva-2022-languages}. In this work, we will focus on Pali written in the Sinhala script (the Sinhala script is also used for writing Pali and Sanskrit literature in Sri Lanka~\cite{gair1996sinhala}), rather than the more commonly known Devanagari script, which presents additional challenges.

The digitisation of historical texts presents several significant challenges. These include the difficulty of OCR on degraded copies, the inconsistency of spelling and orthography over the centuries~\cite{jayatilleke2025sidiac}, and the limited research that has been conducted on OCR for scanned documents and images featuring text written in the Sinhala script~\cite{desilva2026survey}.



We introduce \textsc{SiPaKosa}\footnote{\scriptsize \url{https://github.com/RanxduG/SiPaKosa-Dataset}}, a large-scale corpus combining classical Sinhala Buddhist texts from historical archives with complete canonical scriptures from web sources.
Our contributions include: (1) a corpus of 786,839 sentences ($\sim$9.25M words) from 16 historical documents and complete Tripi\d{t}aka texts, (2) comprehensive metadata and language classification, (4) language-separated subcorpora Sinhala (465,539 sentences, 5.42M tokens), Pali (495 sentences, 3.2K tokens), and Mixed Sinhala-Pali (320,805 sentences, 3.83M tokens), and (5) evaluations showing 3-6 times gaps between proprietary and open-source models in terms of perplexity scores.

\section{Related Work}
\label{sec:related-work}

\subsection{The Bible in 100 languages}

A multilingual Bible Corpus was introduced by~\citet{christodouloupoulos2015massively}, comprising translations into 100 languages.
The corpus was constructed by web scraping publicly available Bible translations, automatic sentence alignment using the inherent verse-level structure of Biblical texts, and finally, quality filtering based on translation completeness and linguistic coverage.
The resulting corpus structure preserves both the original hierarchical metadata (book names, chapter numbers, verse identifiers) and linguistic annotations (sentence boundaries, tokenisation), facilitating diverse research applications from machine translation to typological studies.

While serving different theological traditions, both the Bible and the Buddhist corpora share critical characteristics: these texts hold canonical status within their respective religious traditions (the Bible for Christianity, the Tripiṭaka for Buddhism). This status necessitates not only linguistic accuracy but also the preservation of religious meanings and theological concepts during digitisation. Furthermore, both face challenges related to translation and OCR across diverse linguistic families, along with the responsibility to safeguard sacred texts through digital means while ensuring accuracy for researchers who will utilise this data to study doctrinal texts.

\subsection{Sinhala Diachronic Corpus}

The \textit{Sinhala Diachronic Corpus Version-1.0} (\texttt{SiDiaC-v.1.0}) was introduced by \citet{jayatilleke2025sidiac} and comprises 58,027 word tokens drawn from 46 literary works spanning the 5th to 20th century~CE (426~CE to 1944~CE), sourced from the \textit{National Library of Sri Lanka}.
Text extraction was carried out using \texttt{Google Document AI} OCR, achieving an average accuracy of 96.84\% across all processed documents.
The corpus was carefully constructed to ensure balanced temporal coverage whilst prioritising canonical literary works; among the 46 books included, 18 are religious texts predominantly focused on Buddhism, annotated with a two-layer genre scheme (Fiction/Non-Fiction at primary level; Religious, History, Poetry, Language, and Medical at secondary level).
\texttt{SiDiaC-v.1.0} implements copyright filtering based on Sri Lanka's 70-year post-mortem \textit{auctoris} rule under the Intellectual Property Act No.~36 of 2003, a practice directly adopted by \textsc{SiPaKosa}.
 
The second version of this corpus, \texttt{SiDiaC-v.2.0} \citep{jayatilleke2026sidiacv2}, substantially expands the resource to 241k words across 185 literary works, with publication dates ranging from 1800~CE to 1955~CE and written dates spanning the 5th to the 20th century~CE.
A date-annotated subset of 59 documents totalling approximately 67k words is further stratified by estimated written date, enabling fine-grained diachronic analysis.
Of the 185 documents, 86 are classified under the religious genre, reflecting the enduring centrality of Buddhism to Sinhala literary culture.
\texttt{SiDiaC-v.2.0} addresses several limitations of its predecessor, including more comprehensive post-processing to correct malformed tokens, code-mixed content (Pali, Sanskrit, and English), and multi-column rendering errors introduced by OCR; it also broadens the written-date annotation methodology from strict manuscript-composition dating to the publication-year approach used in corpora such as \texttt{COHA} \citep{davies2012coha}, enabling inclusion of a wider range of literary works.
 
\texttt{SiDiaC} focuses on diachronic language change across secular literary traditions and provides the most directly comparable Sinhala corpus to \textsc{SiPaKosa} in terms of construction methodology.


\begin{table*}[h!tb]
\centering
\small
\setlength{\tabcolsep}{4pt}
\resizebox{\textwidth}{!}{
\begin{tabular}{lllrrl l}
\hline
\textbf{Corpus} & \textbf{Language(s)} & \textbf{Tradition} & \textbf{Tokens} & \textbf{Units} & \textbf{Annotation} & \textbf{Alignment} \\
\hline
\texttt{Bible 100} \citeyearpar{christodouloupoulos2015massively} & 100 langs & Christian & varies & $\sim$31K verses/lang & None & Verse \\
\texttt{CBETA} \citeyearpar{cbeta2023} & Classical Chinese & Buddhist & 100M+ chars & 5,320+ texts & TEI XML & - \\
\texttt{GRETIL} \citeyearpar{tokunaga2009gretil} & 10+ Indic & Multi & GBs (raw) & 1,000+ texts & Raw text & - \\
\textit{Quranic Arabic} \citeyearpar{dukes2010morphological} & Arabic & Islamic & 77,430 words & 6,236 \={a}yahs & Morph+Syn+Sem & Verse \\
\texttt{Tanzil} \citeyearpar{zarrabi2014tanzil} & 40+ langs & Islamic & varies & 6,236 verses/lang & None & Verse \\
\texttt{DCS} \citeyearpar{hellwig2010dcs} & Sanskrit & Multi & 2.5M+ items & $\sim$650K sents & Lemma+POS+Morph & - \\
\texttt{SansTib} \citeyearpar{nehrdich2022sanstib} & Skt-Tibetan & Buddhist & 14.4M tokens & 317K pairs & Aligned & Sentence \\
\texttt{MITRA} \citeyearpar{nehrdich2026mitra} & Skt+Zh+Tib+Pali & Buddhist & varies & 1.74M pairs & MT-aligned & Sentence \\
\texttt{SiDiaC-v.2.0} \citeyearpar{jayatilleke2026sidiacv2} & Sinhala & Mixed & 241K words & 185 works & Genre+date & - \\
\textsc{SiPaKosa} [our work] & Sinhala + Pali & Buddhist & \textbf{9.25M tokens} & \textbf{786K sents} & Lang. class + meta & - \\
\hline
\end{tabular}}
\caption{Comparison of \textsc{SiPaKosa} with related religious and Indic text corpora.}
\label{tab:related_work_comparison}
\end{table*}

\subsection{Religious and Cultural Text Corpora}

The digitisation of religious texts in low-resource languages presents unique challenges that require not only linguistic accuracy but also cultural sensitivity and the preservation of specialised terminology.
Several major religious corpora initiatives have established important precedents across traditions.
 
The Digital Pali Canon provides comprehensive coverage of Therav\={a}da scriptures in Pali,\footnote{\scriptsize \url{https://tripitaka.online/}} whilst the \textit{BDK English Tripi\d{t}aka}\footnote{\scriptsize \url{https://www.bdkamerica.org/tripitaka-list/}} offers scholarly English translations of the Chinese Tripi\d{t}aka.
The \textit{Chinese Buddhist Electronic Text Association} (\texttt{CBETA}) represents one of the most comprehensive Buddhist digitisation efforts globally, providing over 100 million characters of Chinese Buddhist texts drawn from the Taish\={o} Shinsh\={u} Daiz\={o}ky\={o} (Volumes~1-55 and 85; 5,320+ individual texts) and supplementary collections \citep{cbeta2023}.
Files are encoded in \texttt{TEI$\sim$P5 XML} with Unicode handling for 2,800+ rare Buddhist characters proposed to the Unicode Consortium, released under \texttt{CC BY-NC-SA 2.5$\sim$TW}.
 
The \texttt{GRETIL} (\textit{G\"{o}ttingen Register of Electronic Texts in Indian Languages}) project provides over 1,000 electronic editions primarily in Sanskrit, with additional texts in Pali, Prakrit, Tamil, Malayalam, Hindi, Marathi, Old Javanese, and Tibetan \citep{tokunaga2009gretil}.
The total collection spans several gigabytes in plain text, \texttt{HTML}, and \texttt{TEI-conformant XML}, released under \texttt{CC$\sim$BY} via \texttt{Zenodo}.
However, \texttt{GRETIL} provides minimal annotation (raw text only) and an ongoing migration from legacy encodings; whilst it includes some Pali texts, it provides no NLP-ready annotation layers, unlike \textsc{SiPaKosa}.
 
The \textit{Quranic Arabic Corpus} \citep{dukes2010morphological} provides morphological annotation of all 77,430 words of the Qur'an across 128,219 morphemic forms (clitic-level segmentation) using a 44-tag POS tagset derived from traditional Arabic grammar, spanning 114 surahs and 6,236 \={a}yahs.
The corpus provides four annotation layers: morphological segmentation, part-of-speech tagging, syntactic dependency analysis, and semantic ontology, making it the most methodologically comparable sacred-text annotation project to what \textsc{SiPaKosa} could aspire to in future extensions.
Building upon this foundation, the \textit{Quranic Arabic Dependency Treebank} (\texttt{QADT}) \cite{dukes2013syntactic} extends morphological analysis with gold-standard syntactic annotation for approximately 37,578 words ($\sim$49\% of the Qur'an), achieving an F-measure of 78\% with a rule-based dependency parser trained on the annotated data.
 
The \textit{Tanzil parallel corpus} \citep{zarrabi2014tanzil} offers Qur'anic translations in over 40 languages with verse-level alignment across 6,236 verses and 114 surahs, facilitating cross-lingual Islamic studies.
The \textit{Hebrew Bible Treebank} \citep{sichel2002hebrew} offers morphological and syntactic annotation of Biblical Hebrew with detailed morphological features following traditional Hebrew grammar.
The \textit{Digital Corpus of Sanskrit} (\texttt{DCS}) \citep{hellwig2010dcs} contains approximately 650,000 sentences with over 2.5 million lexical items annotated for lemmatisation, POS tagging, and morphological analysis, covering Sanskrit texts from 500~BCE to 1900~CE; later extensions added syntactic annotation in Universal Dependencies format \citep{hellwig2020dcs} and word sense annotation.
At 650k sentences, the \texttt{DCS} is the most directly comparable single-language annotated ancient-text corpus to \textsc{SiPaKosa}'s 786k sentences, and both address low-resource ancient languages; however, \texttt{DCS} does not cover Buddhist Pali in the Sinhala script, nor does it address language mixing between a living vernacular and a classical liturgical language.
 
These projects collectively demonstrate shared requirements: (1)~domain expertise for doctrinal accuracy, (2)~preservation of specialised religious terminology, (3)~careful handling of language mixing between sacred languages (e.g.\ Pali-Sinhala, Arabic-Urdu, Sanskrit-Hindi), and (4)~maintenance of cultural context. Table \ref{tab:related_work_comparison} presents a systematic comparison of \textsc{SiPaKosa} against related religious and Indic text corpora across six dimensions: language coverage,  religious tradition, scale, annotation depth, and alignment granularity.

\section{Methodology}
\label{sec:methodology}

\begin{figure}[t]
\centering
\includegraphics[width=0.95\columnwidth]{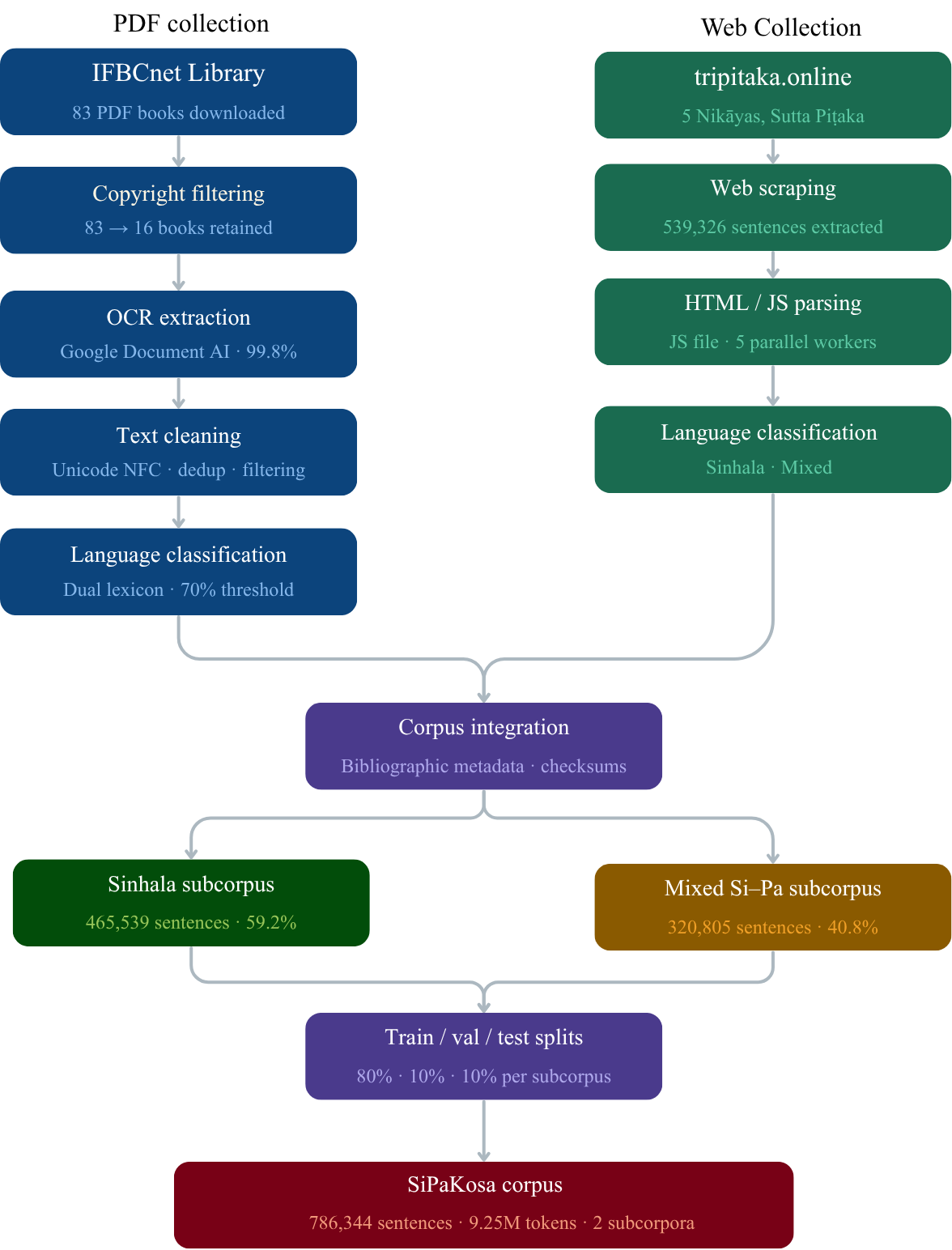}
\caption{Complete methodology pipeline showing dual-source data collection, processing, and integration.}
\label{fig:methodology}
\end{figure}

We employ a dual-source methodology in order to cover canonical Buddhist scriptures with historically archived texts to achieve comprehensive coverage of Sinhala and Pali Buddhist literature. Our complete pipeline from data source identification through final corpus creation is illustrated in Figure~\ref{fig:methodology}. Furthermore, the book level filtering from a total of 83 to 16 eligible books is depicted in Figure~\ref{fig:sankey}. 

\subsection{Data Collection}

Our historical corpus is compiled from the \texttt{IFBCnet Library}\footnote{\scriptsize \url{https://download.ifbcnet.org/}}, a comprehensive digital archive of Sinhala and Pali Buddhist texts along with web-scrapping \texttt{tripitaka.online}\footnote{\scriptsize \url{https://tripitaka.online/}}, to complement historical texts with canonical scriptures.
We downloaded 83 PDF books from \texttt{IFBCnet Library}, representing the complete available collection, each accompanied by bibliographic metadata including title, author information, publication year, and category classifications.
The archive spans publications from the early days through the mid-20th century, providing valuable historical coverage of Sinhala and Pali Buddhist scholarship during a critical period of Buddhist revival and modernisation in Sri Lanka.

To ensure legal compliance, we implemented rigorous copyright filtering based on Sri Lanka's 70-year post-mortem auctoris rule under the Intellectual Property Act No. 36 of 2003\footnote{\scriptsize \url{https://www.gov.lk/wordpress/wp-content/uploads/2015/03/IntellectualPropertyActNo.36of2003Sectionsr.pdf}} following the methodology by~\citet{jayatilleke2026sidiacv2}.
Firstly, Tripiṭaka canonical texts in PDF format were excluded via hardcoded rules, as these would be acquired separately through web scraping to ensure canonical completeness.
Secondly, books authored by individuals who passed away before 1954 were classified as public domain with high confidence.
Finally, books with living authors or incomplete author metadata were conservatively excluded to minimise legal risk.
This conservative approach prioritises legal certainty over corpus size.

\begin{figure}[t]
\centering
\includegraphics[width=\columnwidth]{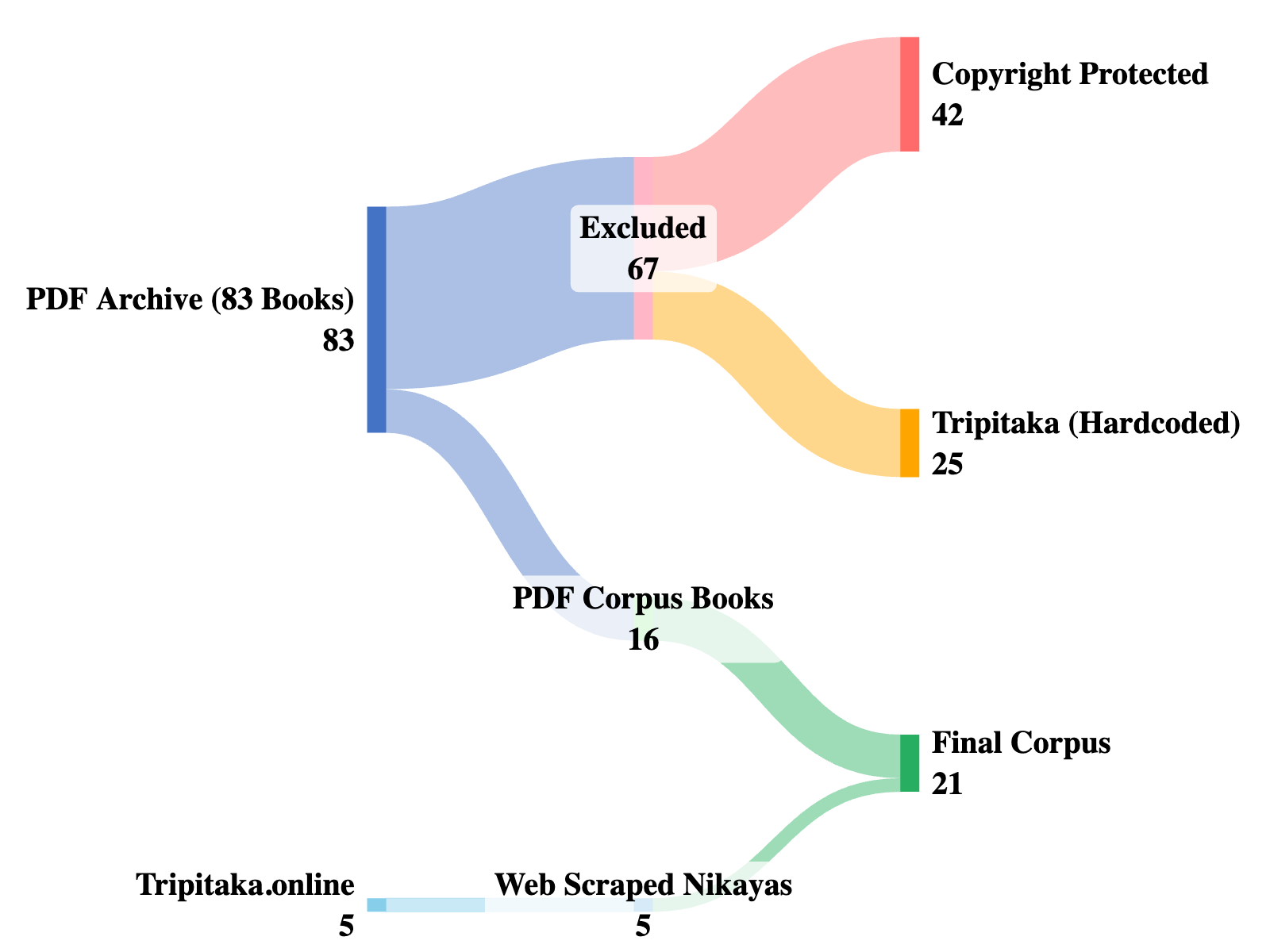}
\caption{Book-level filtering from 83 downloaded books to 16 corpus-eligible books.}
\label{fig:sankey}
\end{figure}

After copyright filtering, 16 books (19.3\% of downloaded collection) were deemed suitable for inclusion.
These books span three traditional Sinhala and Pali Buddhist literary categories: ``Books related to Tipitaka'' comprising 5 books (31.25\%), which include canonical translations and commentaries; ``Old Buddhist books'' comprising 8 books (50.0\%), which encompass philosophical treatises, devotional literature, and scholarly works; and ``Buddhist characters'' comprising 3 books (18.75\%), which consist of hagiographies and biographical accounts of prominent Buddhist figures. The entire filtering process, from initial collection to the selection of books for the final corpus, is illustrated in Figure~\ref{fig:sankey}.

In addition to the historical PDF corpus, we systematically web-scraped \texttt{tripitaka.online}, a comprehensive digital repository of the Theravada Pali Canon maintained by Buddhist scholars, which provides Sinhala-language translations of the canonical scriptures alongside the original Pali text. The Canon, known as the Tripi\d{t}aka, literally \textit{three baskets} in Pali, comprises the Vinaya Pi\d{t}aka (monastic discipline), the Sutta Pi\d{t}aka (discourses of the Buddha), and the Abhidhamma Pi\d{t}aka (systematic philosophical analysis).
Our scraping focused on the five Nikayas of the Sutta Pitaka: Dīgha Nikāya (Long Discourses, 34 suttas), Majjhima Nikāya (Middle-Length Discourses, 152 suttas), Saṃyutta Nikāya (Connected Discourses, 65 suttas), Aṅguttara Nikāya (Numerical Discourses, 1,365 suttas), and Khuddaka Nikāya (Minor Collection, 831 suttas).
Each sutta on tripitaka.online presents Pali canonical verses alongside corresponding Sinhala translations, providing natural parallel text alignment that preserves the semantic correspondence between source and target languages (this corpus lacks clear parallel text alignment but contains useful information for translation work). The detailed implementation is provided in Appendix~\ref{app:scraping}.

\subsection{Text Extraction and Processing}
\label{subsec:data-extraction}

We developed a two-stage extraction pipeline combining traditional PDF text extraction with OCR-based processing to handle the diverse quality of historical scans.
For each PDF, we first attempted direct text extraction using \texttt{pdfplumber}, a Python library optimised for extracting text from PDF documents. As it did not produce precise extractions, we employed \texttt{Google Document AI}\footnote{\scriptsize \url{https://cloud.google.com/document-ai/}}~\cite{jayatilleke-de-silva-2025-zero}, a production-grade OCR service with demonstrated strong performance on Indic scripts, including Sinhala~\cite{jayatilleke2025sidiac}.

This way, we ensure that if we can collect data from the PDF, we can get it without major issues or inconsistencies, but if the page is a scanned image, we have to use the alternative route of OCR extraction.

For pages processed through \texttt{Document AI}, we implemented comprehensive quality control measures based on character-level confidence scores returned by the OCR engine.
The OCR process achieved an average character-level confidence of 99.8\% across all processed pages, demonstrating both the quality of the source scans and \texttt{Document AI}'s effectiveness on Sinhala script.
Pages were classified into two confidence tiers: high confidence (mean confidence >=0.85, representing 85\% of OCR-processed pages) and low confidence (mean confidence < 0.85, representing 15\% of pages).
Low-confidence pages were flagged for potential manual review and excluded from the primary corpus but preserved for future improvement efforts.

The proportion of pages achieving high confidence (85\%) demonstrates both the quality of the archive scans and \texttt{Document AI}'s effectiveness on Sinhala script.
We also implemented automated page classification to identify and filter non-content pages, such as covers, title pages, tables of contents, indices, and blank pages, using pattern-matching rules, retaining 7,064 content pages (99.8\% retention).

For web-scraped canonical texts, we implemented a systematic crawler to extract structured \texttt{HTML} content from \texttt{tripitaka.online}, whilst preserving both Pali canonical text and Sinhala translations separately. After careful analysis, we identified that the data displayed on \texttt{tripitaka.online}, comes from a \texttt{js} (\texttt{JavaScript}) file, so we made sure the web scraper algorithm took this \texttt{js} file and read its content to get the textual data we needed, we preserved structural metadata during this process to organise the whole process which included nikaya identifiers, book numbers, sutta identifiers, and verse numbers for precise alignment and reference purposes.
The extraction process required 72 hours, using multi-threading with 5 parallel workers and rate-limiting constraints to ensure server stability, reflecting the substantial scale of the canonical corpus (539,326 sentences across five Nikayas).

\subsection{Language Classification}
\label{subsec:lang-class}

Buddhist texts in Sri Lanka traditionally mix Sinhala and Pali with varying proportions depending on text type, historical period, and intended audience.
To facilitate both monolingual and cross-lingual research, we implemented an automated language classification mechanism using lexicon-based matching.
We constructed two primary lexicons: a Sinhala lexicon based on \textit{Google's open-source Sinhala Pronunciation Lexicon}\footnote{\scriptsize \url{https://raw.githubusercontent.com/google/language-resources/master/si/data/lexicon.tsv}} containing 41,617 word forms covering a vast vocabulary, and a custom Pali lexicon containing 14,278 terms extracted from a Sinhala-Pali dictionary, which was included as one of the 16 books and then filtered for uniqueness to avoid overlap with Sinhala words. Further information on constructing the custom Pali Lexicon can be found at Appendix \ref{app:pali-lexicon}

Pages and sentences were classified using a combined confidence scoring system that integrates lexicon-based word matching with morphological pattern analysis.
For each text segment, we calculated a weighted confidence score for both Sinhala and Pali, where lexicon coverage (proportion of words matching each language's lexicon) received 70\% weight and morphological feature detection (case markers, verbal endings, particles) received 30\% weight.
Text was classified as Sinhala only if the Sinhala confidence score reached at least 70\% and exceeded the Pali confidence score.
Text was classified as Pali if the Pali confidence score reached at least 70\% and exceeded the Sinhala confidence score.
Text was classified as Mixed if neither language's confidence score reached the 70\% threshold, indicating substantial presence of both languages or ambiguous linguistic features.
The 70\% threshold was selected empirically after manual inspection of sample texts to balance precision (avoiding false classifications) and recall (capturing genuine monolingual content).

\subsection{Corpus Integration and Metadata}

We combined sentences from both sources into unified language-based subcorpora designed to support diverse research applications.
The Sinhala subcorpus combines historical PDF Sinhala pages (111,632 sentences) with Tripitaka Sinhala translations (353,907 sentences), totalling 465,539 sentences representing purely Sinhala content suitable for monolingual language modelling and Sinhala-specific NLP tasks.
The Mixed subcorpus combines historical PDF mixed pages (135,386 sentences) with Tripitaka Pali canonical text (185,419 sentences), totalling 320,805 sentences in both Sinhala and Pali that capture the language-mixing characteristic of Buddhist scholarly discourse.

For each book in the corpus, we collected comprehensive metadata in accordance with best practices in digital humanities corpus construction \cite{jayatilleke2025sidiac}. The schema captures five categories of information: bibliographic details (source URLs, Sinhala and English titles, publication dates, and publisher information); author records (full names, dates of birth and death, and biographical notes); technical provenance (file sizes, SHA-256 checksums, download timestamps, and OCR confidence statistics); classification labels (category assignments, language classification results, and historical period); and copyright status (public domain eligibility and confidence level of assessment). The full metadata schema is provided in Figure~\ref{fig:metadata-schema}.

We created train-validation-test splits at 80-10-10 ratios for each language subcorpus.
For the Sinhala subcorpus, this resulted in 372,431 sentences for training, 46,553 sentences for validation, and 46,555 sentences for testing.
For the Mixed subcorpus, this yielded 256,644 for training, 32,080 sentences for validation, and 32,081 sentences for testing.
The training set comprises 629,075 sentences (80\% of the total corpus), the validation set comprises 78,633 sentences (10\%), and the test set comprises 78,636 sentences (10\%) when aggregated across all language subcorpora and converted to sentence-level statistics.
These splits are fixed and are released with the corpus to enable direct comparison of future research results.


\section{Evaluation of \textsc{SiPaKosa}}

\subsection{Corpus Statistics}
\label{sec:corpus-statistics}

\textsc{SiPaKosa} comprises 786,839 total sentences from dual complementary sources: 247,513 sentences extracted from 16 copyright-cleared historical PDF books spanning 7,064 content pages, and 539,326 sentences from five Nikayas of web-scraped canonical texts. A comprehensive statistical analysis of the complete integrated corpus is presented in Table~\ref{tab:corpus-stats}.

\begin{table}[!htb]
\centering
\resizebox{\columnwidth}{!}{
\begin{tabular}{l|l|r}
\hline
\textbf{Source} & \textbf{Metric} & \textbf{Count} \\
\hline
\multirow{3}{*}{Historical PDF Corpus} & Documents (books) & 16 \\ \cline{2-3}
& Content pages & 7,064 \\ \cline{2-3}
& Sentences & 247,513 \\ 
\hline
\multirow{2}{*}{Web-Scraped Canonical Texts} & Nikayas covered & 5 \\ \cline{2-3}
& Sentences & 539,326 \\
\hline
\multirow{3}{*}{\textbf{Total}} & \textbf{Sentences} & \textbf{786,344} \\ \cline{2-3}
& \textbf{Tokens} & \textbf{9,249,792} \\ \cline{2-3}
& \textbf{Average tokens/sentence} & \textbf{11.76} \\
\hline
\end{tabular}}
\caption{Overall corpus statistics combining historical and canonical texts.}
\label{tab:corpus-stats}
\end{table}

\subsubsection{Language Distribution}

The Sinhala subcorpus dominates at 59.2\% of the total corpus (465,539 sentences), reflecting our focus on Sinhala Buddhist literature.
The mixed subcorpus comprises 40.8\% (320,805 sentences), capturing canonical Pali scriptures and mixed commentarial traditions. The statistics of each language subcorpus, revealing the complementary contributions of our dual sources, are detailed in Table~\ref{tab:language-distribution}.

\begin{table}[h!tb]
\begin{center}
\resizebox{0.98\columnwidth}{!}{
\begin{tabular}{l|l|c|c}
\hline
\textbf{Type} & \textbf{Metric} & \textbf{Sinhala} & \textbf{Mixed} \\
\hline
\multirow{4}{*}{Sentences} &
PDF corpus & 111,632 & 135,386 \\ \cline{2-4}
& Web-scraped & 353,907 & 185,419 \\ \cline{2-4}
& \textbf{Total Count} & \textbf{465,539} & \textbf{320,805} \\ \cline{2-4}
& \% of corpus & 59.2\% & 40.8\% \\
\hline
\multirow{2}{*}{Tokens} &
\textbf{Total Count} & \textbf{5,418,716} & \textbf{3,831,076} \\ \cline{2-4}
& \% of corpus & 58.6\% & 41.4\% \\
\hline
\end{tabular}}
\caption{Sentence and token distribution by language subcorpus and source.}
\label{tab:language-distribution}
\end{center}
\end{table}
                        
\begin{figure}[h!tb]
\begin{center}
\includegraphics[width=\columnwidth]{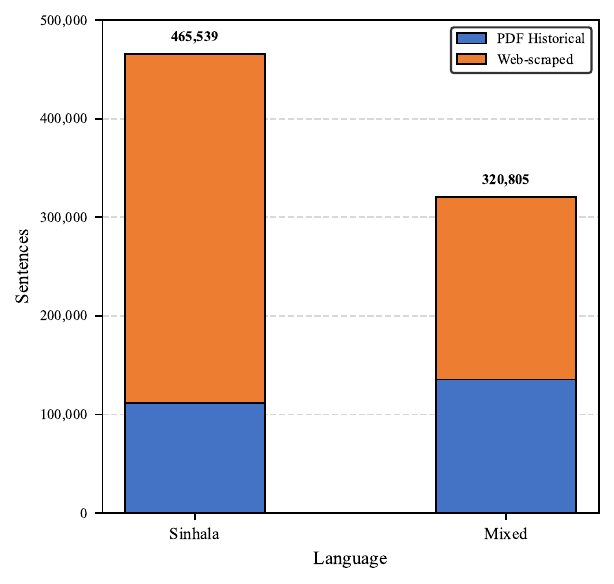}
\caption{Sentence distribution by language and source.}

\label{fig:language-distribution-chart}
\end{center}
\end{figure}

The Sinhala subcorpus (465,539 total sentences) is predominantly web-scraped (76\% from Tripitaka, 24\% from PDF), reflecting the extensive online availability of canonical translations. The Mixed subcorpus (320,805 total) shows more balanced contributions (58\% web-scraped, 42\% PDF), capturing both canonical Pali verses and historical commentarial discourse.
Based on our methodology for language classification in \ref{subsec:lang-class}, there was a possibility of 3 language classes forming; unfortunately, no Pali-only content was found. The dual-source composition of each language subcorpus is illustrated in Figure~\ref{fig:language-distribution-chart}.

At the token level, the corpus comprises 9,253,009 tokens, with the Sinhala subcorpus containing 5,418,716 tokens (58.6\%) and the Mixed subcorpus containing 3,831,076 tokens (41.4\%). The average sentence length varies across subcorpora: 11.6 tokens per sentence for Sinhala, 11.9 tokens per sentence for Mixed text.

\subsection{Results}
\subsubsection{Evaluation Setup}
\begin{table*}[!htb]
\begin{center}
\setlength{\arrayrulewidth}{0.8pt}
\small
\resizebox{0.85\textwidth}{!}{
\begin{tabular}{l|r|r|r|r|r}
\hline
\multirow{2}{*}{\textbf{Model}} & \textbf{Buddhist} & \textbf{Mixed} & \textbf{General} & \textbf{BS/G} & \textbf{M/G} \\
 & \textbf{Sinhala} & \textbf{Si-Pa} & \textbf{Sinhala} & \textbf{Ratio} & \textbf{Ratio} \\
\hline
\texttt{GPT-3.5-Turbo} & \textbf{1.09} & \textbf{1.11} & \textbf{1.08} & 1.01 & 1.03 \\
\texttt{GPT-4o-mini} & 1.23 & 1.15 & 1.31 & 0.94 & 0.88 \\
\texttt{GPT-4-Turbo} & 1.56 & 1.59 & 1.48 & 1.05 & 1.07 \\
\texttt{GPT-4o} & 1.62 & 1.77 & 1.68 & 0.96 & 1.06 \\
\hline
\texttt{Llama-3.1-8B-Instruct} & 3.29 & 4.18 & 3.26 & 1.01 & 1.28 \\
\texttt{Aya-Expanse-8B} & 6.21 & 8.82 & 4.90 & 1.27 & 1.80 \\
\texttt{Llama-3.2-3B-Instruct} & 6.62 & 7.81 & 6.94 & 0.95 & 1.13 \\
\texttt{Qwen2.5-3B-Instruct} & 7.24 & 9.81 & 6.56 & 1.10 & 1.49 \\
\texttt{Gemma-2-9B-It} & 22.21 & 36.71 & 14.85 & 1.50 & 2.47 \\
\texttt{SinLlama} & 183.33 & 189.67 & 101.57 & 1.80 & 1.87 \\
\hline
\end{tabular}}
\caption{Comparison of perplexity among proprietary language models and small language models using different corpora.}
\label{tab:perplexity-main}
\end{center}
\end{table*}
We evaluate corpus quality and domain characteristics using nine state-of-the-art language models selected to represent diverse architectures, parameter scales, and training paradigms.
Proprietary models include models from the \texttt{GPT} family; \texttt{GPT-3.5-Turbo}, \texttt{GPT-4o-mini}, \texttt{GPT-4-Turbo}, and \texttt{GPT-4o} \cite{openai2023gpt4}.
Open-source models include \texttt{Llama-3.1-8B-Instruct} and \texttt{Llama-3.2-3B-Instruct} \cite{dubey2024llama}, \texttt{aya-expanse-8b} \cite{aryabumi2024aya}, \texttt{Qwen2.5-3B-Instruct} \cite{qwen2024qwen25}, and \texttt{Gemma-2-9B-it} \cite{gemmateam2024gemma2} with explicit multilingual and Sinhala support.

We evaluate models on three carefully constructed test sets designed to assess different aspects of Sinhala language modelling capability.
The Buddhist Sinhala test set comprises 1,024 sentences sampled using diversity sampling from the pure Sinhala test split.
The Mixed Sinhala-Pali test set comprises 1,024 sentences from the mixed test split, representing mixed Buddhist scholarly discourse.
The General Sinhala test set comprises 1,024 sentences from a CulturaX\footnote{\scriptsize \url{https://huggingface.co/datasets/uonlp/CulturaX}}, which consists of Sinhala news articles and online content.
To ensure representative coverage while avoiding redundancy, we employ diversity sampling using K-means clustering (k=4) on multilingual sentence embeddings, selecting sentences closest to cluster centroids to capture diverse linguistic phenomena within each corpus.

We report corpus-level perplexity calculated as the exponentiated average negative log-likelihood across all tokens in each test set.
For open-source models with direct access to model internals, given a test set with $N$ tokens $w_1, w_2, \ldots, w_N$, perplexity is computed as;

\begin{equation}
\mathbf{PPL} = \exp\left(-\frac{1}{N}\sum_{i=1}^{N} \log P(w_i | w_{<i})\right)
\end{equation}

where $P(w_i | w_{<i})$ is the model's predicted probability for token $w_i$ given preceding context.
For proprietary models that do not expose direct token-level log probabilities, we employ an approximate evaluation methodology in which models assess linguistic quality per sentence, with response confidence serving as a proxy for model certainty about the input text.
Detailed proprietary model evaluation methodology is provided in Appendix~\ref{app:api-perplexity}. To quantify domain-specific challenges independent of absolute perplexity values, we compute domain gap ratios as;

\begin{equation}
    \mathbf{BS/G} = \frac{\mathrm{PPL}_{\mathrm{Buddhist-Sinhala}}}{\mathrm{PPL}_{\mathrm{General}}}
\end{equation}

\begin{equation}
    \mathbf{M/G} = \frac{\mathrm{PPL}_{\mathrm{Mixed}}}{\mathrm{PPL}_{\mathrm{General}}}
\end{equation}

\label{sec:results}

A comprehensive perplexity results across all nine models and three test corpora are presented in Table~\ref{tab:perplexity-main}.
The results reveal substantial performance variation between proprietary models and open-source alternatives.

\subsubsection{Model Performance Analysis}

\texttt{GPT-3.5-Turbo} achieves the best overall performance with the lowest perplexities of 1.09 for Buddhist Sinhala, 1.11 for Mixed Sinhala-Pali, and 1.08 for General Sinhala, demonstrating strong Sinhala language comprehension.
All four proprietary models substantially outperform all locally-hosted open-source models, with \texttt{GPT-3.5-Turbo}, \texttt{GPT-4o-mini}, \texttt{GPT-4-Turbo}, and GPT-4o achieving perplexities between 1.08 and 1.77 across all test sets.

Among open-source models, \texttt{Llama-3.2-3B-Instruct} leads with perplexities of 3.29 for Buddhist and 4.18 for Mixed corpora, representing the best available option for resource-constrained settings.
\texttt{Aya-Expanse-8B}, despite explicit multilingual training including Sinhala, achieves moderate performance (6.21 Buddhist, 8.82 Mixed), whilst smaller 3B models (\texttt{Llama-3.2-3B-Instruct, Qwen2.5-3B-Instruct}) demonstrate competent performance.
\texttt{Gemma-2-9B-It} significantly underperforms with perplexities exceeding 14 on all corpora despite its 9B parameter count, indicating fundamentally inadequate Sinhala coverage in pretraining data.

Domain gap ratios reveal how models handle the specialised vocabulary and discourse patterns of Buddhist literature compared to general Sinhala.
\texttt{GPT-3.5-Turbo} exhibits minimal domain gap with a BS/G ratio of 1.01 and M/G ratio of 1.03, suggesting a promising generalisation across registers and domains.
\texttt{GPT-4o-mini} exhibits an inverse domain gap on Buddhist texts (BS/G ratio of 0.94), actually performing better on classical Buddhist Sinhala than general text.
\texttt{Llama-3.1-8B} shows generalisation on pure Sinhala (BS/G ratio 1.01) but moderate difficulty with mixed text (M/G ratio 1.28).
\texttt{Aya-Expanse-8B} and \texttt{Gemma-2-9B} exhibit larger domain gaps (1.27-2.47), with Gemma's extreme M/G ratio of 2.47 indicating the comprehension difficulty with language mixing.

Mixed Sinhala-Pali corpora present substantially elevated perplexity across most models, with an average M/G ratio of 1.42 across all models excluding the \texttt{Gemma} outlier.
This consistent difficulty possibly reflects fundamental challenges in handling intra-sentential language mixing, in which Pali canonical quotations and Sinhala explanatory prose alternate rapidly.
Proprietary models handle language mixing most effectively with M/G ratios ranging from 0.88 to 1.07.
Open-source models struggle more significantly: \texttt{Llama-3.1} shows 28\% perplexity increase (M/G ratio 1.28), \texttt{Aya-Expanse} shows 80\% increase (1.80), and \texttt{Qwen2.5} shows 49\% increase (1.49).

Another finding of this evaluation is the substantial performance gap between proprietary models and locally hosted open-source alternatives.
\texttt{GPT-3.5-Turbo} outperforms the best local model \texttt{Llama-3.1-8B} by a factor of 3.0 on Buddhist Sinhala (perplexity 1.09 vs 3.29).
This gap persists across all three test sets, suggesting it reflects fundamental differences in the scale and quality of the training data.
The gap has critical practical implications: whilst proprietary models offer exceptional performance for production Buddhist NLP applications, they incur per-token costs and lack transparency.
Open-source alternatives enable offline deployment, fine-tuning for specific Buddhist NLP tasks, and cost-effective experimentation.

Our findings for open source models challenge the common belief that larger models always outperform smaller ones, showing more complex relationships between parameter count and performance in low-resource languages.
\texttt{Llama-3.2-3B} (perplexity 6.62) outperforms the larger \texttt{Aya-Expanse-8B} (6.21) despite having fewer than half the parameters, whilst \texttt{Gemma-2-9B}'s catastrophic performance (22.21) demonstrates that scale alone provides no guarantees.
For proprietary models where accuracy is paramount, and API costs are acceptable, \texttt{GPT-3.5-Turbo} and \texttt{GPT-4o-mini} provide powerful performance.
For research systems requiring local deployment and customisation, \texttt{Llama-3.1-8B} represents the best foundation despite its 3 times higher perplexity.
In resource-constrained settings, \texttt{Llama-3.2-3B} or \texttt{Qwen2.5-3B} offer strong performance with manageable computational requirements.

\section{Conclusion}
\label{sec:conclusion}

We have presented \textsc{SiPaKosa}, a comprehensive corpus of canonical Sinhala and Pali Buddhist texts compiled from dual sources: historical public domain archives and systematically web-scraped canonical scriptures.
Through rigorous digitisation, OCR processing, web scraping, copyright filtering, and quality assurance, we created a high-quality resource comprising 786,839 sentences (9.25 million tokens) across Sinhala, Pali, and mixed subcorpora, addressing critical gaps in Sinhala NLP while preserving important cultural and religious heritage.

Our comprehensive baseline evaluations across nine language models demonstrate significant performance variation, with proprietary models achieving remarkably low perplexities of 1.08 to 1.77 and substantially outperforming open-source alternatives with perplexities of 3.29 to 36.71.
This performance gap highlights both the current state of Sinhala language modelling and opportunities for improvement in open-source multilingual systems.
All models struggle with Sinhala-Pali mixed text, indicating opportunities for specialised model development targeting classical Buddhist literature.

This corpus serves as a critical resource for diachronic linguistic studies, providing insights into the history and evolution of language, as well as the profound influence of religious thought on everyday speech. In addition to formal canonical analysis, this corpus allows examination of folk speech and literary proverbs derived from classical Buddhist texts~\cite{sofalas-etal-2026-sinfos}. This reflects the historical process by which scriptural concepts were integrated into the broader Sinhala linguistic identity.

\section*{Limitations}
\label{sec:limitations}

\paragraph{Temporal and Genre Coverage:}
Copyright restrictions exclude scholarly works published after the 1950s, potentially missing modern interpretations and contemporary Buddhist scholarship.

\noindent
\paragraph{Post-Processing:}
Although the OCR extraction pipeline achieved an average character-level confidence of 99.8\% across all processed pages, no manual post-processing was conducted on the extracted text. Correcting residual OCR errors and normalising historical  spelling variations in classical Sinhala requires deep linguistic expertise in  diachronic Sinhala and P\={a}li that was not available to the authors;  consequently, such errors and orthographic inconsistencies may persist in the corpus.

\noindent
\paragraph{Language Classification:}
Buddhist texts frequently mix Sinhala and Pali within sentences, making strict language separation imperfect.
The Mixed category encompasses diverse text types, from balanced bilingual to predominantly Pali canonical texts.
The 70\% classification threshold is somewhat arbitrary and may benefit from optimisation. Additionally, this study did not explore potential Sanskrit text in these documents due to the absence of advanced language classification models. 

\noindent
\paragraph{Evaluation Limitations:}
Standard perplexity does not directly measure downstream task performance or semantic understanding.
We lack human assessments of model outputs on Buddhist question answering or text generation tasks.
Cannot analyse internal mechanisms of proprietary models or verify their Sinhala training data.
Could evaluate other proprietary models rather than only OpenAI models

\section{Bibliographical References}
\label{main:ref}
\bibliographystyle{lrec2026-natbib}
\bibliography{lrec2026-example}

\appendix

\section{Additional Related Work}
\label{app:religious_corpora}

\subsection{Multilingual Evaluation Benchmarks}
\label{app:benchmarks}
 
Multilingual evaluation frameworks include XGLUE \citep{liang2020xglue} (11 NLU and NLG tasks across 19 languages) and XTREME-R \citep{ruder2021xtremer} (10 tasks across 50 languages) for cross-lingual NLU tasks, building on English-only benchmarks GLUE \citep{wang2018glue} (9 NLU tasks, English only), SuperGLUE \citep{wang2019superglue} (8 NLU tasks, English only), and SQuAD \citep{rajpurkar2016squad} (100,000+ extractive QA pairs, English only).
 
Language-specific MMLU variants have proliferated, covering Arabic \citep{koto2024arabicmmlu}, Chinese \citep{li2024cmmlu}, Turkish \citep{yuksel2024turkishmmlu}, Indonesian \citep{koto2023indommlu}, Korean \citep{son-etal-2025-kmmlu}.
Low-resource language benchmarks include AfriMMLU \citep{adelani2025afrimmlu} for African languages, MalayMMLU \citep{poh2024malaymmlu}, BertaQA for Basque \citep{etxaniz2024bertaqa}, and MILU for 11 Indic languages \citep{verma2025milu}.
The Include benchmark \citep{singhG2025globalmmlu} attempts comprehensive multilingual coverage, though many benchmarks rely on automatic translation that can introduce errors and fail to capture cultural context \citep{ji2023cultural}.
 
Beyond the major religious corpus projects discussed in Sections~2.3-2.5, several specialised initiatives contribute to computational religious studies.
 
The Quranic PropBank \citep{zaghouani2012quranic} extends semantic role labelling to Classical Arabic religious texts.
 
\textbf{Christian Resources.} The Semantic Bible project provides multilingual semantic role annotation for Biblical texts with ontological relationships \citep{semanticbible2015}.
 
\textbf{Hindu and Vedic Resources.} The Sansknet\footnote{\scriptsize \url{https://www.wilbourhall.org/sansknet/}} project provides computational resources for Sanskrit including morphological analysers adapted to Vedic grammar.
 
\textbf{Multi-Traditional Resources.} The Sacred Texts Archive \citep{sacredtexts2020} provides digitised versions of religious texts across 12 major traditions, though without linguistic annotation.
 
 
 
 
 
\subsection{Indic Language Resources}
\label{app:indic}
 
\textsc{SiPaKosa} exists within a vibrant Indic NLP ecosystem.
Samanantar \citep{ramesh2022samanantar} is the largest publicly available Indic parallel corpus, comprising 49.7 million sentence pairs across 11 Indic languages; however, Pali is entirely absent from the collection, and the general-domain web-crawled text differs fundamentally from the specialised Buddhist canonical register that \textsc{SiPaKosa} targets.
IndicCorp~v2 \citep{doddapaneni2023indiccorpv2} provides 20.9 billion tokens across 24 Indic languages with the IndicXTREME benchmark (105 evaluation sets, 9 tasks); it includes Sinhala but not Pali, and its web-crawled sources contain no Buddhist domain content.
The AI4Bharat-IndicNLP Corpus \citep{kunchukuttan2020ai4bharat} established 2.7 billion words for 10 Indic languages with pre-trained embeddings, providing foundational resources for the broader Indic NLP ecosystem.
 
These resources collectively provide broad coverage of general-domain Sinhala web text but offer zero coverage of Pali and zero domain-specific Buddhist text, confirming that \textsc{SiPaKosa} addresses a genuine and unmet need within the Indic NLP landscape.
 
\subsection{Language Identification}
\label{app:langid}
 
Since Sinhala and Pali share the same script (the Sinhala script is also used for writing Pali and Sanskrit in Sri Lanka; \citealt{gair1996sinhala}), language identification is a core technical challenge for \textsc{SiPaKosa}.
\citet{barman2014code} established that approximately 7\% of tokens in code-mixed text are ambiguous, a finding relevant to the Mixed Sinhala-Pali subcorpus.
 
\subsection{Information Retrieval}
\label{app:ir}
 
Dense passage retrieval \citep{karpukhin2020dense} establishes neural retrieval baselines for open-domain question answering.
For low-resource languages, optimal transport distillation \citep{huang2023improving} and zero-shot linguistic similarity transfer \citep{chari2025improving} improve cross-lingual IR effectiveness.
Structured retrieval frameworks \citep{lee2024tree} provide QA methodologies that could be adapted for Sinhala Buddhist question answering.

\subsection{Domain Adaptation}
Parameter-efficient fine-tuning approaches, including LoRA \citep{hu2021lora} and QLoRA \citep{dettmers2023qlora}, enable domain adaptation without full fine-tuning, while knowledge distillation \citep{gou2021knowledge} supports creating smaller domain-specific models.
These techniques are directly applicable to adapting general-purpose Sinhala models such as \texttt{SinBERT} \citep{dhananjaya2022sinbert} and \texttt{SinLlama} \citep{aravinda2025sinllama} for Buddhist domain tasks using \textsc{SiPaKosa} as the training corpus.
 
\subsection{Cross-Lingual Transfer}
\label{app:xling}
 
\citet{garciaferrero2025crosslingual} provide comprehensive analysis of cross-lingual transfer techniques, whilst TransTokenization \citep{remy2024transtokenization} demonstrates vocabulary initialisation benefits for low-resource language adaptation of LLMs.
\citet{hangya2022improving} shows strategies for improving low-resource languages in multilingual models through vocabulary adaptation and continued pretraining, findings directly applicable to extending multilingual models for Sinhala Buddhist text.
\subsection{LLM Evaluation} 
LLM evaluation surveys \citep{chang2023survey,adlakha2024evaluating} discuss correctness and faithfulness assessment in instruction-following models, informing our baseline evaluation methodology.
\citet{hutchinson-2024-modeling} specifically addresses ethical and methodological considerations when using religious texts in NLP, including data provenance, cultural contexts, and proselytism concerns; this work directly informs \textsc{SiPaKosa}'s commitment to copyright compliance and respectful representation of Buddhist canonical materials.

\section{Additional Methodology}

\subsection{Web-scrapping}
\label{app:scraping}

The web crawler for \texttt{tripitaka.online} implemented the following technical specifications:
Rate limiting was set to 2 requests per second with exponential backoff retry logic (initial delay: 1s, maximum delay: 60s) to handle transient network failures.
The crawler navigated the four-level hierarchy (Nikaya → Book → Sutta → Verse), extracted both Pali canonical text and Sinhala translations separately from the \texttt{JavaScript} file and preserved structural metadata, including nikaya identifiers, book numbers, sutta identifiers, and verse numbers for precise alignment.
We employed a multi-threaded strategy with 5 parallel workers scraping 5 different suttas simultaneously, reducing total extraction time from an estimated 360 hours (sequential processing) to 72 hours (parallel processing), a 5× speedup.
However, even with this optimisation, the extraction required 3 days of continuous operation due to the corpus scale (539,326 sentences) and necessary rate limiting (2 requests per second with exponential backoff) to avoid overwhelming the source server.

\subsection{PDF Processing Flowchart}
\begin{figure}[h!tb]
\begin{center}
\includegraphics[width=0.99\columnwidth]{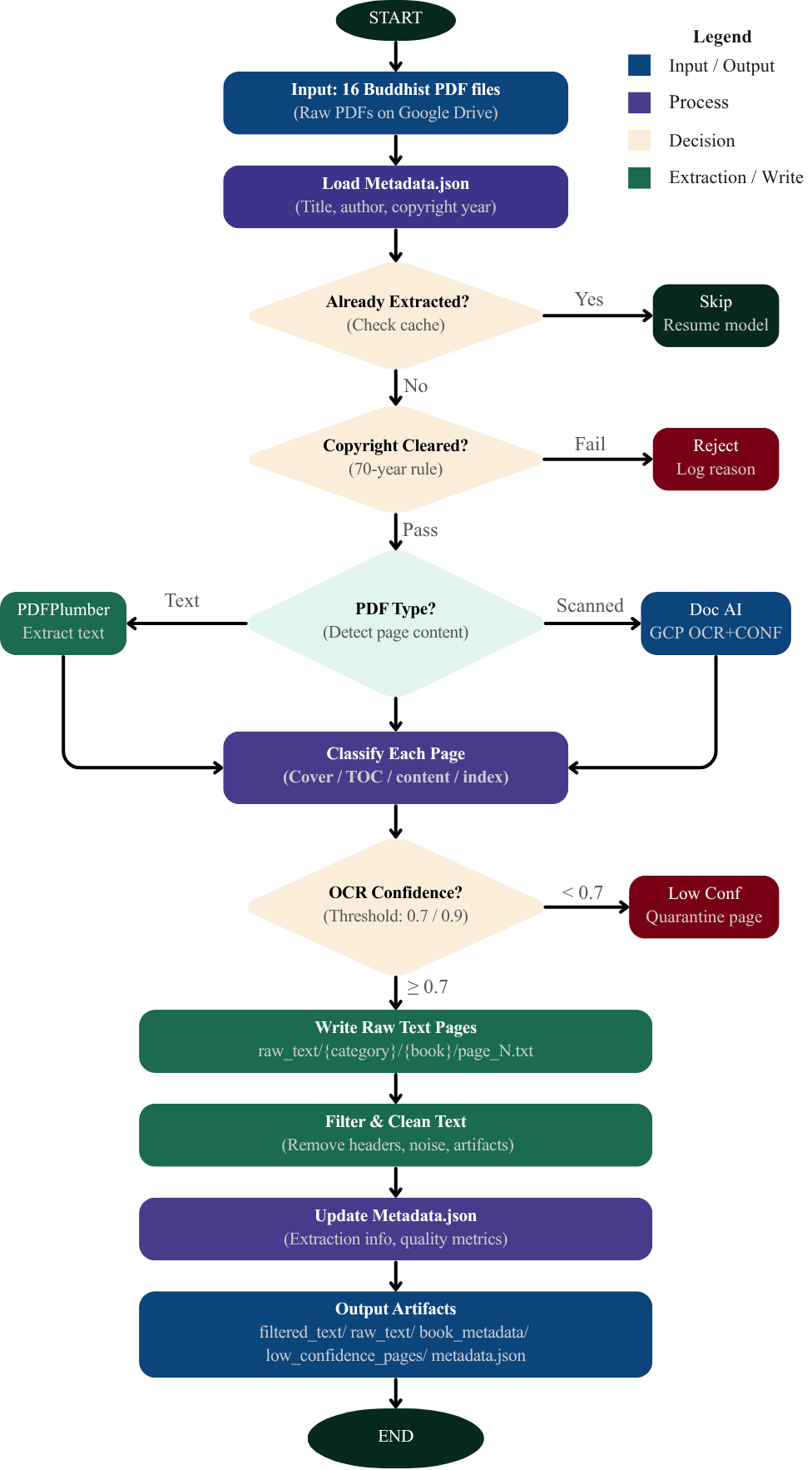}
\caption{PDF processing pipeline applied to the 16 copyright-cleared historical Buddhist books, combining \texttt{pdfplumber} text extraction and \texttt{Google Document AI OCR}, with copyright checking, page classification, and confidence-based quality filtering (threshold: 0.7) prior to corpus integration.}
\label{fig:language-distribution}
\end{center}
\end{figure}

Figure~\ref{fig:language-distribution} illustrates the end-to-end extraction pipeline applied to each of the 16 copyright-cleared PDF books. The pipeline begins by checking whether a book has already been processed, allowing interrupted runs to resume without reprocessing. Books that pass the 70-year post-mortem copyright check are then routed through one of two extraction paths depending on page content: digitally typed pages are handled by \texttt{pdfplumber}, while scanned pages are passed to \texttt{Google Document AI}, which returns character-level OCR confidence scores alongside the extracted text. All pages, regardless of extraction route, are classified into one of four structural categories: cover, table of contents, content, or index, and pages falling below the 0.7 confidence threshold are quarantined rather than discarded, preserving them for potential future review. Content pages that pass the confidence gate are written to disk as raw text, filtered to remove headers and OCR artefacts, and their provenance and quality metrics are recorded back into the per-book metadata record. The final output for each book comprises four artefact directories and an updated \texttt{metadata.json} file.

\subsection{Custom Pali Lexicon}
\label{app:pali-lexicon}

No publicly available Pali lexicon suitable for language classification existed at the time of this research, necessitating the construction of a custom resource. The source material was a Sinhala-Pali dictionary recovered as one of the 16 copyright-cleared books retained in the extraction pipeline described in Section~\ref{subsec:data-extraction} (Refer Table \ref{tab:corpus-books}). All headword entries and their
corresponding Pali equivalents were extracted from the dictionary's structured text output produced by the OCR stage, yielding a raw lexicon of 14,663 unique word forms.
To isolate vocabulary that is diagnostically Pali rather than shared with modern Sinhala, the raw lexicon was filtered against a 41,617-word Sinhala lexicon obtained from Google's open-source language resources repository.
Any token present in the Sinhala lexicon was removed from the Pali word
list, on the basis that a word attested in the standard Sinhala vocabulary
is not reliably diagnostic of Pali in a language classification context.
The overlap analysis identified 887 shared words (6.05\% of the raw Pali
lexicon), predominantly short cognates of three to seven characters that
are common to both languages due to their shared Indo-Aryan heritage.
After removing these overlapping entries, the final filtered Pali lexicon
comprised 13,776 unique word forms. This lexicon was then used alongside
the Sinhala lexicon in the dual-lexicon classifier described in the
following section to assign language labels at the paragraph and page
level.

\subsection{Corpus Metadata Information}

\begin{figure}[H]
\centering
\includegraphics[width=0.5\columnwidth]{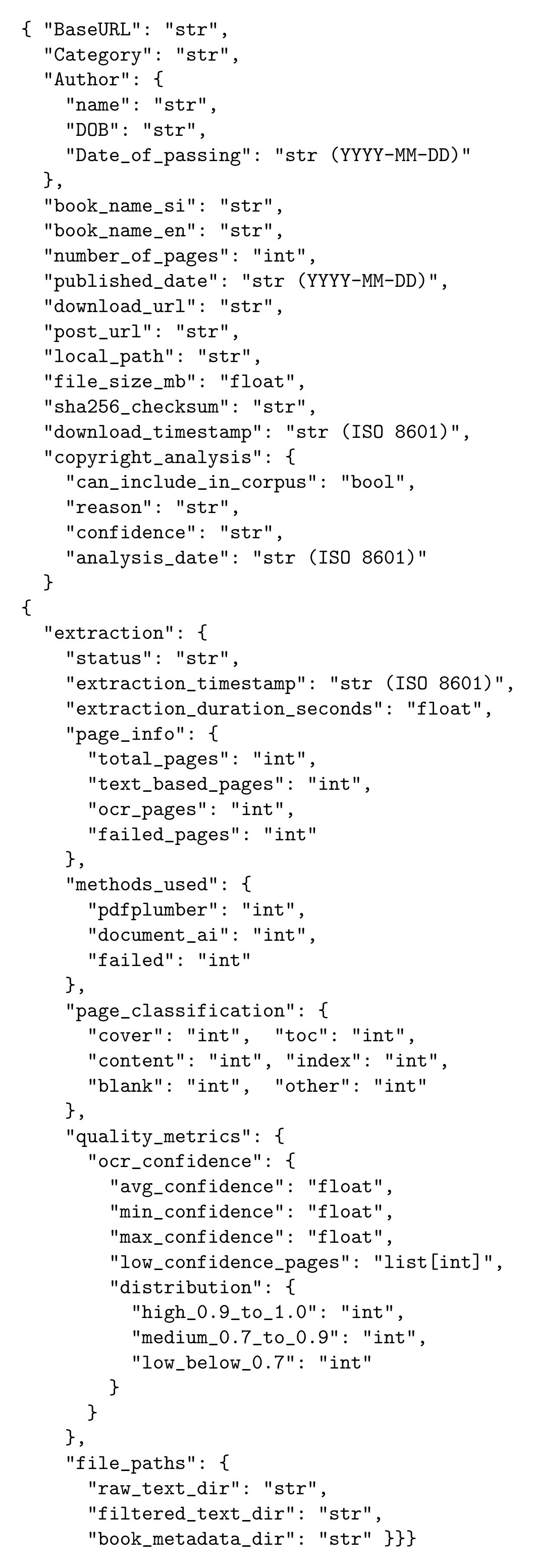}
\caption{JSON structure of a metadata record of the books in the IFBC corpus, defining the fields and their corresponding data types.}
\label{fig:metadata-schema}
\end{figure}

\subsection{Copyright Filtered Books}
\label{app:book-data}

The list of copyright-filtered books is listed in Table~\ref{tab:corpus-books}.

\begin{table*}[!htb]
\begin{center}
\small
\resizebox{\textwidth}{!}{
\begin{tabular}{lllc}
\toprule
\textbf{No.} & \textbf{Title} & \textbf{Author} & \textbf{Published} \\
\midrule
\multicolumn{4}{c}{\textbf{Books Related to Tipitaka (5 books)}} \\\hline
1 & Maithree Buddha Wanshaya nohoth anagatha wanshaya & Ven. Wilgammula Sangharaja Thero & 2015 \\
2 & Milind Prashna & Ven. Heenatikubure Sumangala Nahimi & 2015 \\
3 & Pretha Wasthu & Dhammapala Thero & 2015 \\
4 & Vimana Wasthu & Dhammapala Thero & 2015 \\
5 & Wishuddhi Margaya & Buddhaghosa Maha Thero & 2015 \\
\midrule
\multicolumn{4}{c}{\textbf{Old Buddhist Books (8 books)}} \\\hline
6 & Jathila Thotilla & S. Mahinda Thero & 2015 \\
7 & Pancha Maha Wadaya & Ven. Migettuwatte Gunananda Thero & 2015 \\
8 & Pali Bhasha Shabdakoshaya & Ven. Widurupola Piyatissa Mahanayaka Thero & 2015 \\
9 & Sinhala Deepavansaya & Unknown & 2015 \\
10 & Buthsarana & Vidyachakrawarthi & 2015 \\
11 & The Book of Dalada Pujavaliya & Unknown & 2015 \\
12 & The Mahavamsa & Ven. Hikkaduwe Sri Sumangala Ther & 2015 \\
13 & Saddharma Ratnavali & D.B. Jayatilaka & 2015 \\
\midrule
\multicolumn{4}{c}{\textbf{Buddhist Characters (3 books)}} \\\hline
14 & Ven. Kadawadduwe Jinalankara Thero's Character & IFBC & 2015 \\
15 & Ven. Man Buridaththa Thero's Character & IFBC & 2015 \\
16 & Ven. Rerukane Chandawimala Maha Nayaka Thero's Character & IFBC & 2015 \\
\bottomrule
\end{tabular}}
\caption{Copyright-cleared historical Buddhist texts included in the corpus (n=16). All books are in the public domain, with authors who passed away before 1954 (70+ years post-mortem).}
\label{tab:corpus-books}
\end{center}
\end{table*}

\section{Additional Result Information}
\subsection{Proprietary Model Evaluation}
\label{app:api-perplexity}

Proprietary models present a fundamental challenge for perplexity evaluation: production APIs typically do not expose token-level log probabilities required for exact perplexity calculation.
Fortunately, OpenAI's API provides access to token-level logprobs through the \texttt{logprobs} parameter, enabling precise perplexity measurement for all four evaluated GPT variants.

\subsubsection{OpenAI API Logprobs Implementation}

For \texttt{GPT-3.5-Turbo}, \texttt{GPT-4o-mini}, \texttt{GPT-4-Turbo}, and \texttt{GPT-4o}, we utilise OpenAI's \texttt{logprobs} API parameter to extract token-level log probabilities for each input sentence.
The evaluation process proceeds as follows:

\begin{enumerate}
\item For each test sentence, we make an API call with \texttt{logprobs=True} and \texttt{top\_logprobs=1} to request token-level probability information.
\item The API returns log probabilities for each token in the input sequence (via the \texttt{prompt\_logprobs} field in API responses where available, or through the \texttt{logprobs.content} structure in newer API versions).
\item We extract the log probability $\log P(w_i | w_{<i})$ for each token $w_i$ in the sequence.
\item Perplexity is computed using the standard formula: $\text{PPL} = \exp\left(-\frac{1}{N}\sum_{i=1}^{N} \log P(w_i | w_{<i})\right)$.
\item We implement rate limiting (1-2 second delay between requests) to comply with API usage guidelines and avoid rate limit errors.
\end{enumerate}

This methodology provides exact perplexity measurements for OpenAI models, identical in principle to the evaluation of open-source models with direct access to model internals.
The primary difference is computational: API-based evaluation incurs per-token costs and requires network latency for each request, whilst local evaluation provides faster throughput at the cost of computational infrastructure requirements.

\subsubsection{Model-Specific Considerations}

\textbf{\texttt{GPT-3.5-Turbo}:}
This provides the baseline for OpenAI's Sinhala capabilities.
The API consistently provides logprobs access, enabling reliable perplexity measurement.
\\
\textbf{\texttt{GPT-4o-mini}:}
This lightweight variant of \texttt{GPT-4o} is optimised for cost-efficiency whilst maintaining strong multilingual capabilities.
Logprobs access is consistent across API versions.
\\
\textbf{\texttt{GPT-4-Turbo}:}
The speed-optimised variant of \texttt{GPT-4} provides full logprobs access.
Note that ``Turbo'' designation indicates architectural optimisations for inference speed rather than model capacity changes.
\\
\textbf{\texttt{GPT-4o}:}
The ``omni'' variant represents OpenAI's latest multimodal architecture with enhanced multilingual capabilities.
Full logprobs access is provided, though API response structure differs slightly from earlier \texttt{GPT-4} variants (using \texttt{logprobs.content} rather than \texttt{prompt\_logprobs}).

\subsection{Proprietary vs Open-Source Models}

The perplexity values obtained through OpenAI's API are directly comparable to those from open-source models evaluated locally, as both methods compute the identical mathematical quantity: the exponentiated average negative log-likelihood.
The key advantages of API-based evaluation include:
\begin{itemize}
\item No local computational infrastructure required
\item Access to proprietary models unavailable for local deployment
\item Consistent evaluation environment (model version, ttokenisation inference configuration)
\end{itemize}

The primary disadvantages include:
\begin{itemize}
\item Per-token API costs accumulating across 1,024 sentences × 3 corpora
\item Network latency and rate limiting, extending evaluation time
\item Dependency on API availability and version stability
\item Limited transparency regarding model architecture and training data
\end{itemize}


\end{document}